# DAGAM: Data Augmentation with Generation And Modification

**Byeong-Cheol Jo[1], Tak-Sung Heo[1], Yeongjoon Park[1]**
**Yongmin Yoo[1], Won Ik Cho[2], Kyungsun Kim[1]**

AI R&D Group, NHN Diquest[1]
Department of Electrical and Computer Engineering and INMC, Seoul National University[2]
{byeongcheol7674, gjxkrtjd221, yeongjoon1227, yooyongmin91}@gmail.com
tsatsuki@snu.ac.kr, kksun@diquest.com

**Abstract**

Text classification is a representative downstream task of natural language processing, and has exhibited excellent performance since the advent of pre-trained language models based on Transformer architecture. However, in pre-trained language models, under-fitting often occurs due to the size of the model being very large compared to the amount of available training data. Along with significant importance of data collection in modern machine learning paradigm, studies have been actively conducted for natural language data augmentation. In light of this, we introduce three data augmentation schemes that help reduce underfitting problems of large-scale language models. Primarily we use a generation model for data augmentation, which is defined as Data Augmentation with Generation (DAG). Next, we augment data using text modification techniques such as corruption and word order change (Data Augmentation with Modification, DAM). Finally, we propose Data Augmentation with Generation And Modification (DAGAM), which combines DAG and DAM techniques for a boosted performance. We conduct data augmentation for six benchmark datasets of text classification task, and verify the usefulness of DAG, DAM, and DAGAM through BERT-based fine-tuning and evaluation, deriving better results compared to the performance with original datasets.

## 1. Introduction

Text classification is a representative downstream task of natural language processing (NLP), and studies in various domains are being actively conducted. The text classification task are in relevance with several domains such as intention classification, topic classification, sentiment analysis, etc. (Jang et al., 2019; Kim and Jeong, 2019; Risch and Krestel, 2019; Li et al., 2020; Heo et al., 2021). Since the advent of pre-trained language models (PLMs) such as Bidirectional encoder representations from transformers (BERT), Transformer-based deep learning models have exhibited excellent performance for text classification (Vaswani et al., 2017; Yu et al., 2019; Devlin et al., 2019; Guo et al., 2020; Shaheen et al., 2020).

However, Transforemer-based deep learning models may yield underfitting because the size of the model can be too extensive compared to the size of the training data (Liu et al., 2019). In this regard, some studies have reported that the performance can be improved in various tasks by artificially increasing the size of the data (Liu et al., 2019; Brown et al., 2020). In data-driven machine learning, collecting sufficient amount of high-quality data is definitely an important process for an adequate level of model learning, but since such collection processes are not always viable, many studies tackle this issue from the perspective of augmentation using pre-existing data (Yu et al., 2018; Wei and Zou, 2019; Feng et al., 2019; Shorten and Khoshgoftaar, 2019; Xie et al., 2020; Feng et al., 2020).

Two kinds of strategies are mainly adopted for natural language data augmentation. The first is to collect data using human resources, and the other is to create and modify data mechanically or semi-automatically. The former guarantees data quality, but collecting and pre-processing large-scale data manually is extremely time-consuming and costly. Therefore, various automation strategies were proposed to overcome such limitation. In representative approaches, data is augmented by using a generation model or modifying a part of the text (Yu et al., 2018; Shorten and Khoshgoftaar, 2019; Xie et al., 2020).

As one of the studies using generation models, Yu et al. (2018) proposed a back-translation method using both direction of machine translation systems. Here, the data is augmented by translating an English sentence into French and then translating it back to English again through a French-English translation model. However, since semantic discrepancy can occur in the round-trip-translation to other languages, the augmentation of natural and syntactically plausible sentences is often not guaranteed.

In the approach using text modification, data similar to the original text is augmented by using strategies such as replacing a specific word with a synonym, inserting a random word, changing the position of two random words in a sentence, or deleting a random word (Wei and Zou, 2019). Other studies investigated the effect of giving synthetic noise, replacing words with hyponym and hypernym, and using semantic text exchange (Feng et al., 2019; Feng et al., 2020). However, using a thesaurus such as WordNet or a part-of-speech tagger usually requires considerable amount of time and budget.

To attack the above limitations, we propose a data aug-mentation scheme using a paraphrase-based generation model and character order change (COC) strategy. Our methodology consists of three steps (Figure 1). The first is to augment data using a generation model on raw text, which we define as data augmentation with generation (DAG). The second is to augment data using COC, a strategy that corrupts some words appearing in raw text, and we define this as data augmentation with modification (DAM). Finally, we combine the two methods and call it as data augmentation with generation and modification (DAGAM).

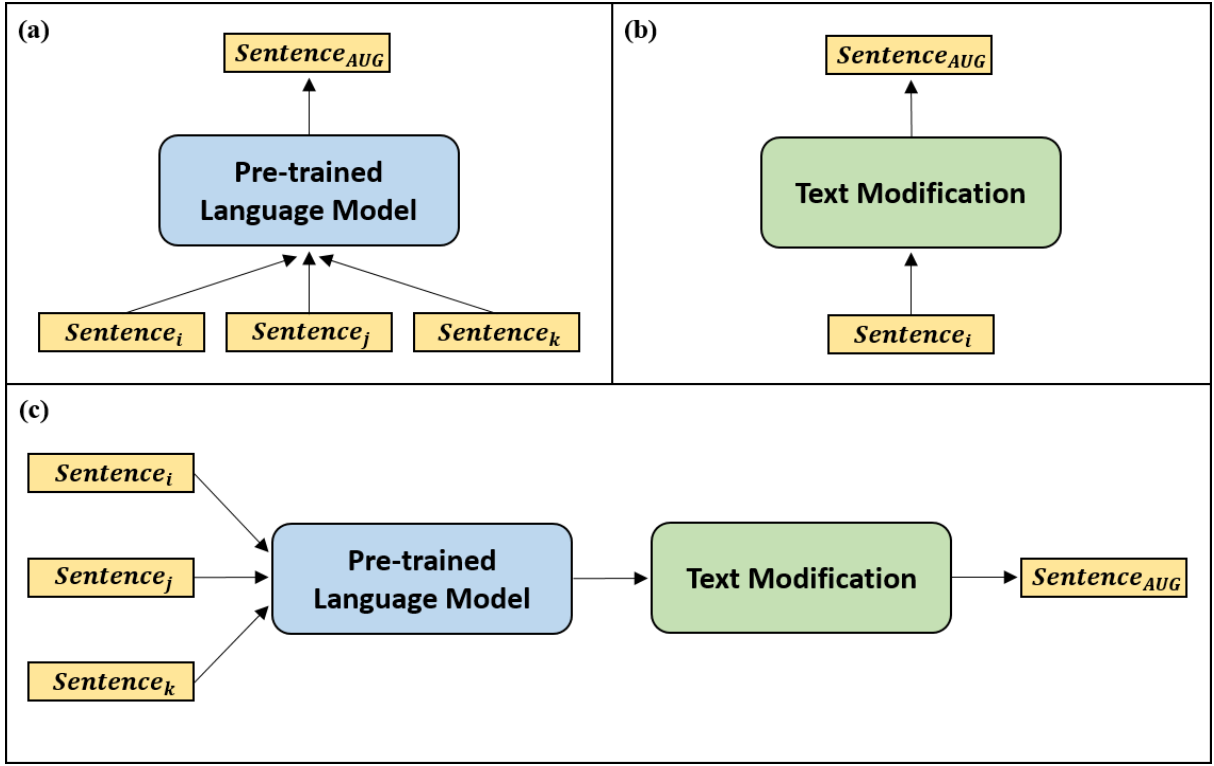

Figure 1: Architecture according to the proposed methods: (a) DAG (b) DAM (c) DAGAM

Our methodology is a simple and easy strategy to automatically augment natural language data. We perform data augmentation on six benchmark datasets in text classification task. To check the power of our scheme, we use BERT, which is a representative Transformer-based pre-trained language model, for the fine-tuning and evaluation. The utility of our methodology is verified by performance improvement made on all of the benchmark datasets, compared to the case no augmentation. The contribution of our work to the field is as follows:

- We propose DAG, a data augmentation method using a generation model, DAM, a data augmentation method using character order changing (COC), and DAGAM, a combined scheme of two methods, verifying and comparing their utility through BERT-based fine-tuning and evaluation.

- We publicly open the codebase for our augmentation experiments, to make it easier for the NLP community of industry and acdemia to easily access data augmentation methodologies.

## 2. Related Work
### 2.1. Data Augmentation Schemes

**Generation models** Recently, various data augmentation studies have adopted trained language generation models. The representative one is Sennrich et al. (2016) which first proposed back-translation approach. Here, sentences of the source language that correspond with the translated sentence of the target language are generated using machine translation models trained with the parallel corpus. In this regard the machine translation performance is improved by simultaneously using the newly created corpus and the existing parallel corpus. This back-translation method is exploited not only in machine translation but also in other NLP fields. Yu et al. (2018) tackles the question answering task, finding an appropriate answer in a document for an input query, with back-translation schemes. Randomly selected sentences in the document are augmented as an additional data after round-trip translation using a pre-trained machine translation model. (Xie et al., 2020) improves the performance of the text classification task by training the classifier in the direction of reducing the distance between the output distribution yielded by the origi- nal sentence and the augmented sentence (generated by back-translation).

**Rule-based augmentation** Rule-based methods that does not exploit generation models are also actively studied in natural language data augmentation. Wei and Zou (2019) improves the performance of text classification tasks by applying a data augmentation method that arbitrarily changes/corrects words or substitutes words using thesaurus. Feng et al. (2019) augments the data with entity replacement using WordNet. Min et al. (2020) performs data augmentation by changing the subject and object in the sentence or using passivization. In addition, Guo et al. (2019) proposes mix-up augmentation, which removes parts of two sentences and then connects them to create a new sentence.

| Type | Sentence |
|---|---|
| Original Text 1 | shell canada said it raised crude prices by canadian cts a barrel today |
| Original Text 2 | conoco raises crude oil prices up to one dlr barrel wti at dlrs |
| Original Text 3 | phillips raises crude postings cts effective today wti to dlrs bbl |
| Summarized Text | shell canada said it raised crude oil prices by canadian cts a barrel. dlrs phillips raised crude oil prices up to one dlr barrel wti to dlrs bbl. |

Table 1: An example of DAG corresponding to crude class of R8.

| Effect | Sentence using Byte Pair Encoding |
|---|---|
| Original | What is a pre ##train ##ed bert model |
| Deletion | What is a parent ##ried bert model |
| Insertion | What is a pre ##train ##ed bert med ##ol |
| Replacement | What is a pre ##train ##ed bret model |

Table 2: The effect of COC when using byte pair encoding.

## 2.2. Pretrained Language Models

**BERT** Before digging into the up-to-date generative models, we think it beneficial to briefly skim BERT (Devlin et al., 2019), a de facto pretrained language model constructed by stacking encoder blocks of Transformer (Vaswani et al., 2017). BERT has already exhibited excellent performance in various NLP fields, and is utilized in the way of i) pre-training with a large-scale corpus and then ii) fine-tuning for down-stream tasks. In pre-training, after randomly masking the tokens in a sequence of sentences, the model parameters are trained with two objectives: the model predicts the original word (masked language model, MLM) and simultaneously predicts the relevance of two given sentences (next sentence prediction, NSP). When fitting on a downstream task, fine-tuning is performed task-wisely on relatively small data in each domain.

**T5** T5 is a model that fully leverages the encoder-decoder structure of Transformer (Raffel et al., 2020). Unlike the BERT model that uses only the encoder block of transformer, T5 consists of an encoder-decoder structure, displaying the characteristics of generative model that all outputs are in the textual format that comes from the decoder. Like BERT, T5 has a pre-training stage and a fine-tuning stage. In the pre-training stage, training encoder is managed by predicting a masked token for each sentence as in BERT. However, in the decoder, after replacing consecutive tokens with a single mask, the model predicts and yields only tokens of the masked part as an output, not the entire input sentence. In the fine-tuning stage, learning is performed for various down-stream tasks such as classification, generation, etc., where the decoder infers the correct answer in the textual format.

## 3. Methodology

We removed the remaining texts except English for the experiment, and to investigate the effect of data augmentation on natural language processing, we propose the following three methods.

### 3.1. Data Augmentation with Generation (DAG)

DAG uses a language generation model for data augmentation. In particular, among generation-based methods such as paraphrasing and summarization, we adopt the latter in our approach. In the case of paraphrasing, output is reasonable for single sentences but not usually when an input is in document-level. In this regard, we summarize a chunk of sentences from the original data and make up a new, longer one, to generate an augmented data that has a similar representation distribution as the original one.

We use a generation model by combining three texts instead of one, to provide variations to input data. We define the summarized text extracted through the generation model as an augmented data, and assign the label corresponding to the original input data as the new target value. Figure 1 (a) describes DAG, where texts $Sentences_i$, $Sentences_j$, $Sentences_k$, sentences with the same label, are used for the data augmentation. An example is shonw in Table 1. We randomly extract three texts of the same class and augment their summarization, at the same time removing duplicate data if exists.

### 3.2. Data Augmentation with Modification (DAM)

Data augmentation with modification (DAM) exploits a psychological phenomenon called the word superiority effect (Coch and Mitra, 2010), and we apply a character order change (COC) strategy, which comes after the phenomenon, to text data. COC denotes fixing the first and the last character in a word and randomly permuting the rest, which brings effects of token insertion, token deletion, token replacement, and anagram, in BERT's byte pair encoding tokenizer as shown in Table 2.

| Sampling Strategy | DAM (n) | DAM (n) | IMDB | AGNews | 20Newsgroup | TREC | R8 | R52 |
|---|---|---|---|---|---|---|---|---|
| TRAIN-ALL | 0 | 0 | 93.65 | 92.8 | 85.17 | 97 | 98 | 95.51 |
|  | 1 | 0 | 93.64 | 92.82 | 85.47 | 97.4 | 98.26 | - |
|  | 0 | 3 | **94.05** | **93.06** | **86.7** | 97 | 98.6 | **97.03** |
|  | 1 | 3 | 93.82 | 92.84 | 86.59 | 97.2 | 98.35 | - |
| TRAIN-HALF | 0 | 0 | 92.82 | 92.18 | 82.74 | 96.6 | 98 | 92.7 |
|  | 1 | 0 | 93.05 | 92.38 | 83.39 | 96.6 | 98.4 | - |
|  | 0 | 5 | 93.31 | 92.52 | 84.82 | 96.4 | 98.6 | 95.78 |
|  | 1 | 5 | 93.4 | 92.44 | 84.8 | **97.4** | **98.67** | - |

Table 3: Experimental results of six benchmark datasets, bolded with the case with the best performance.

For DAM, we first divide the sentence into word units and then randomly extract 20% of tokens. After that, COC is applied to words with a character length of 4 or more (Figure 1, b). We define texts with COC applied as data to be augmented, and removed the duplicated data if exists.

### 3.3. Data Augmentation with Generation And Modification (DAGAM)

In data augmentation with generation and modification (DAGAM), we augment data by combining two strategies proposed in this paper, DAG and DAM. We first obtain summarized data through the generation model, and consequently apply COC thereto. Figure 1 (c) depicts DAGAM, and we removed duplicate data from the data augmented through DAGAM if exists.

## 4. Experiments

The verified the validity of the proposed method through six text classification benchmark datasets, using BERT-based evaluation. In DAG technique, we used T5-base as a generation model, which has shown excellent performance in text summarization area (Raffel et al., 2020).

### 4.1. Benchmark Dataset

We conduct experiments on widely used text classification benchmarks, namely IMDb, AGnews, 20Newsgroups, TREC, R8, and R52.

- IMDb is a binary sentiment analysis task, built upon a movie review dataset.
- Agnews and 20Newsgroups are topic classification tasks made up of news articles.
- TREC is a question classification task, and includes a dataset that aims to classify fact-based questions into broad semantic categories.
- R8 and R52 are topic classification tasks, namely subset datasets of Reuters 21578, a news article.

Table 4 exhibits the specification of benchmark datsets used to check the validity of our scheme, namely the number of classes, the size of train, development, test set.

| Dataset | # Classes | Train set | Dev set | Test set |
|---|---|---|---|---|
| IMDb | 2 | 22,500 | 2,500 | 25,000 |
| AGNews | 4 | 108,000 | 12,000 | 7,600 |
| 20Newsgroup | 20 | 10,163 | 1,130 | 7,528 |
| TREC | 6 | 4,906 | 546 | 500 |
| R8 | 8 | 4,936 | 549 | 2,189 |
| R52 | 52 | 5,878 | 654 | 2,568 |

Table 4: Specification of benchmarks used for validation.

### 4.2. Settings

Our dataset sampling strategy is divided into TRAIN-ALL (using all the volume of a train dataset) and TRAIN-HALF (using a half volume of a train set). For DAG and DAM, "$DAG$ or $DAM = n$" implies that the volume of the data augmented by DAG or DAM equals $n$ times of the volume of the sampled dataset (all or half), where $n \in \{0, 1, 3, 5\}$ (Equations 1a-1d) and $n = 0$ denotes that DAG or DAM is not applied. For R52, DAG and DAGAM were not applied considering some classes with less than three samples.

$$\text{Original} := (DAG = 0) \& (DAM = 0) \quad (1a)$$

$$\text{DAG} := (DAG > 0) \& (DAM = 0) \quad (1b)$$

$$\text{DAM} := (DAG = 0) \& (DAM > 1) \quad (1c)$$

$$\text{DAGAM} := (DAG > 0) \& (DAM > 0) \quad (1d)$$

### 4.3. Results

Experimental results are displayed in Table 3. The number displayed under each dataset means the accuracy, which was obtained by averaging the output of five experiments. We denote `Original` as the case where no augmentation is conducted.

**DAG** In TRAIN-ALL, when DAG is applied, we observed about 0.02%p to 0.4%p performance enhancement in AGNews, 20Newsgroup, TREC, and R8, compared to using `Original`. Also, in TRAIN-HALF, when DAG was applied, models exhibited about 0.2%p to 0.65%p better performance in IMDB, AGNews, 20Newsgroup, and R8 than `Original`. In particular, for DAG in TRAIN-HALF with R8 dataset, the performance recorded about 0.4%p higher even byond the performance of `Origianl` in TRAIN-ALL. Although we could not obtain performance enhancement in some cases, DAG outperforms `Original` in general. From this, it can be inferred that the data generated through DAG is consistent with the original data regarding the distribution of representation.

**DAM** In TRAIN-ALL, when `DAM` is applied, models showed about 0.26%p to 1.53%p better performance compared to `Original`, except for TREC. Also, similarly in TRAIN-HALF, we observed 0.34%p to 3.08%p performance enhancement in all datasets except for TREC. In particular, for R8 and R52 datasets, `DAM` applied in TRAIN-HALF, displayed 0.6%p and 0.27%p higher performance compared to those in TRAIN-ALL. We conclude that token insertion, token deletion, token replacement effects and anagram, which was enabled by DAM using COC, effects the performance of the trained model in a positive way.

**DAGAM** By combining DAG and DAM, we ob- tained performance improvements on all six benchmark datasets. In particular, TREC and R8 showed better performance by 0.4%p or more than when using the original dataset of TRAIN-ALL, even in TRAIN-HALF scenario.

### 4.4. Discussions

Although we verified the effect of the proposed methodology on six benchmark datasets, higher improvement was observed in general with smaller number of training data. This suggests that our strategy can be considered more practical in data shortage scenarios. In addition, this shows that the proposed method meets the necessity of data augmentation and can be usefully utilized when creating a large-scale corpus for language model pretraining for a specific or expert domain.

As a limitation, we observed that the proposed method shows significant performance improvement in datasets such as AGNews and 20Newsgroup, namely topic classication tasks, compared to sentiment analysis (IMDB) or question classification (TREC) task. The topic classification task is generally robust to text combining or word-level perturbation since the term-based approach is usually effective. However, in the case of sentiment analysis, DAG showed low performance enhancement because the summarization can induce modification of sentence semantics by combining sentences with less similarity. Such phenomenon might have been boosted by the task being binary, that the contents among samples of each class was too diverse for the summarization-based augmentation. In addition, in the case of question classification, since the length of the original text is very short and word-level perturbation may cause a shift in the question type, the output of DAM was not consistent with the original dataset, resulting in a marginal enhancement.

We plan to study a data augmentation methodology that more fits with the syntax or semantics of the original data, in order to attack the above limitation while maintaining the performance boost. Since the effectiveness of our approach depends on the characteristics of the downstream task, it seems that the generation methodology should be studied along with the regulation schemes according to the characteristics of each task.

### 5. Conclusion

In this study, we propose three methods to augment natural language data based on existing corpora. The first method is Data Augmentation with Generation (DAG) using a generation model, where sentences belonging to the same label are summarized by a generation model, to be used as an augment the data. The second is Data Augmentation with Modification (DAM) that modifies the existing text by applying COC. Eventually, we augment data using Data Augmentation with Generation And Modification (DAGAM), which is a combination of DAG and DAM.

We applied proposed strategies on six text classification benchmark datasets, and verified the validity of our method through BERT-based evaluation. As a result of the experiment, DAG, DAM, and DAGAM displayed overall performance boost across datasets, showed better results compared to utlizing only the original data. Our results sometimes suggest that generation models and rule-based methods, when used together, can help obtain a significant performance enhancement.

As a future work, we will proceed to tackle the task-specific effectiveness of data augmentation schemes. Our results are to be publicly open for the development of data augmentation research.

### 6. Bibliographical References